\documentclass[letterpaper, 10 pt, conference]{ieeeconf}
 
\let\labelindent\relax

\usepackage{balance} %
\usepackage[font=footnotesize,labelsep=period]{caption}[2022/02/20]
\usepackage{subcaption}
\usepackage{algorithm}
\usepackage{algpseudocode}
\usepackage[table,dvipsnames,x11names]{xcolor}

\newcommand{\PlannerName}{$\texttt{MapExRL}$} %

\IEEEoverridecommandlockouts                              %

\usepackage{graphics} %
\usepackage{epsfig} %
\usepackage{times} %
\usepackage{amsmath} %
\usepackage{amssymb}  %
\usepackage{multirow}
\usepackage{pbox}
\usepackage{amsthm} 
\usepackage{duckuments}
\usepackage{booktabs}
\usepackage{tabularx}
\usepackage{boldline, bbm}
\usepackage{enumitem}

\usepackage{float}
\usepackage{pifont}
\usepackage{cite}
\usepackage{bbm}
\usepackage{graphicx}
\graphicspath{{./figures/}}
\usepackage{multicol}        %
\usepackage{algorithm}
\usepackage{algpseudocode}

\makeatletter
\let\NAT@parse\undefined
\makeatother
\usepackage[
colorlinks,
linkcolor=blue,
citecolor=blue,
filecolor=red,
urlcolor=blue]{hyperref}

\title{\LARGE \bf 
MapExRL: Human-Inspired Indoor Exploration with \\ Predicted Environment Context and Reinforcement Learning %
}

\author{Narek Harutyunyan$^{*1}$, Brady Moon$^{*2}$, Seungchan Kim$^{2}$, Cherie Ho$^{2}$, Adam Hung$^{3}$, Sebastian Scherer$^{2}$
\thanks{* Equal Contributions}%
\thanks{This work was supported by the National Science Foundation Graduate Research Fellowship under Grant No. DGE1745016, a hardware grant from Nvidia, and the Office of Naval Research under Grant No. N00014-21-1-2110 and Contract No. N6833522C0179.
}
\thanks{This work involved human subjects or animals in its research. The authors and CMU's IRB under Application No. STUDY2024\_00000372 confirm that all human/animal subject research procedures and protocols are exempt from review board approval under the 2018 Common Rule 45 CFR 46.104.d.}
\thanks{$^{1}$Author is with the Brown University, Providence, RI, USA {\tt\small narek\_harutyunyan@brown.edu}}%
\thanks{$^{2}$Authors are with the Robotics Institute, School of Computer Science at Carnegie Mellon University, Pittsburgh, PA, USA
{\tt\small \{bradym, seungch2, cherieh, basti\}@andrew.cmu.edu}}%
\thanks{$^{3}$Author is with the University of Michigan, Ann Arbor, MI, USA {\tt\small adamhung@umich.edu}}%
}
\begin{document}

\thispagestyle{empty}
\pagestyle{empty}

\makeatletter
\let\@oldmaketitle\@maketitle
\renewcommand{\@maketitle}{\@oldmaketitle
\centering
\captionsetup{type=figure, singlelinecheck=false}
\noindent\begin{tabular}{@{}l@{}}
\includegraphics[trim={0.5cm 0cm 0cm -0.5cm},clip,width=\textwidth]{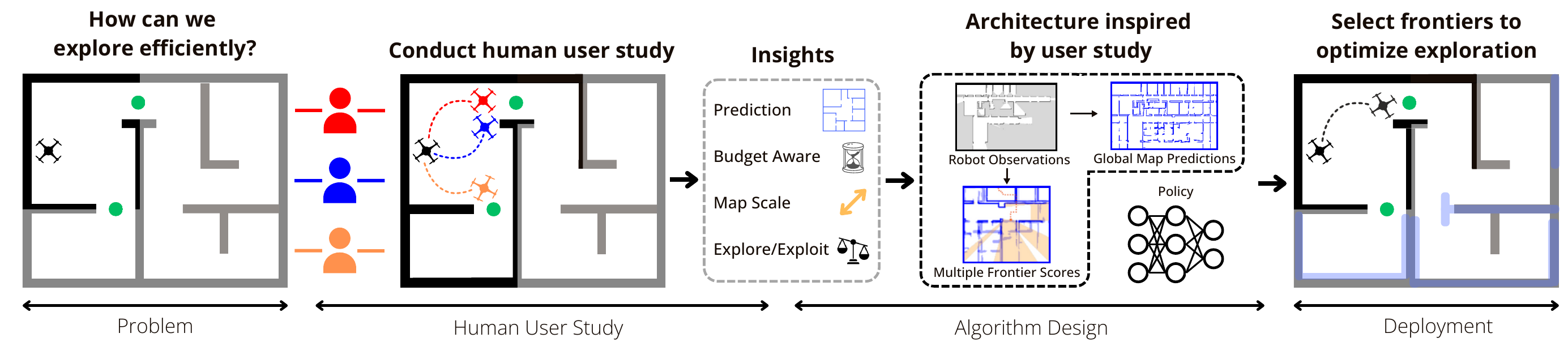}
\end{tabular}
\caption{
An overview of the development and research process for \PlannerName. This work examines the question: How can a robot explore efficiently? We conduct a human user study to gain insights into effective exploration strategies. These insights, in turn, inform the design of our learning-based exploration policy, leveraging global map predictions and other environmental contexts and enabling state-of-the-art performance.
} 
\label{fig:splash}
\setcounter{figure}{1}
}
\makeatother

\maketitle

\begin{abstract}
Path planning for robotic exploration is challenging, requiring reasoning over unknown spaces and anticipating future observations. Efficient exploration requires selecting budget-constrained paths that maximize information gain. Despite advances in autonomous exploration, existing algorithms still fall short of human performance, particularly in structured environments where predictive cues exist but are underutilized. Guided by insights from our user study, we introduce \PlannerName, which improves robot exploration efficiency in structured indoor environments by enabling longer-horizon planning through a learned policy and global map predictions. Unlike many learning-based exploration methods that use motion primitives as the action space, our approach leverages frontiers for more efficient model learning and longer horizon reasoning. Our framework generates global map predictions from the observed map, which our policy utilizes, along with the prediction uncertainty, estimated sensor coverage, frontier distance, and remaining distance budget, to assess the strategic long-term value of frontiers. By leveraging multiple frontier scoring methods and additional context, our policy makes more informed decisions at each stage of the exploration. We evaluate our framework on a real-world indoor map dataset, achieving up to an 18.8\% improvement over the strongest state-of-the-art baseline, with even greater gains compared to conventional frontier-based algorithms. Website: \href{https://mapexrl.github.io}{mapexrl.github.io}
\end{abstract}

\section{Introduction}
Robot exploration is a long-studied problem in robotics due to its wide applications, including search and rescue, reconnaissance, and information gathering.
In exploration, a robot navigates in an unknown space to build map representations and understand the environment.  At its core, it is the problem of deciding where to go to gain the most information \cite{yamauchi1997frontier, oriolo1998real, best2023multi}. A conventional paradigm of deciding where to go is frontier-based exploration \cite{yamauchi1997frontier}, in which a robot incrementally updates the boundary between known and unknown spaces, or \textit{frontier}, and systematically visits the frontiers. While this approach has proven to be robust in large-scale environments \cite{best2023multi}, it heavily relies on simple heuristics, such as visiting the nearest frontiers first \cite{yamauchi1997frontier}, or hand-crafted heuristics \cite{kim2023multi}. However, such algorithms are often short-sighted, reasoning only on the next step. In contrast, humans excel at crafting long-term strategies even with limited observations, that balance efficiency, uncertainty, and resource constraints. \textit{How can we achieve human-level long-horizon exploration?}

To extend the perception horizon in robot exploration, some studies have proposed methods that go beyond reasoning about only the observed areas. These approaches use observed data to predict regions beyond what has been seen \cite{georgakis2022upen, shrestha2019mappred, ho_kim2024mapex}. By leveraging deep learning models, these works generate broader map predictions from partial observations of the environment, improving exploration strategies. However, while these works augment perception by employing predictions, their decision-making remains focused on greedy and myopic action selection.

Reinforcement learning (RL) has been used to train robots to learn navigation policies through trial and error \cite{niroui2019deep, li2019deep}. Previous works have shown that data-driven RL can outperform classical planning approaches \cite{lee2021extendable, cao2023ariadne}. However, these RL-based methods often rely on motion-primitives for action spaces, which could present challenges for long-horizon planning in complex, large-scale environments.

Existing approaches remain limited in achieving long-term human-level decision making for exploration. Toward human-level exploration, we identify two key gaps: (1) an insufficient understanding of \textit{what} signals humans use to make long-horizon decisions and \textit{how} they leverage them, and (2) the lack of a planning framework that integrates these human-inspired insights into robotic exploration.

To bridge these gaps, we conduct a user study comparing human performance to a state-of-the-art (SOTA) exploration algorithm \cite{ho_kim2024mapex} and derive insights to guide our approach. Based on these insights, we introduce \PlannerName, a \textit{budget-aware} RL-based exploration framework that, like humans, leverages predictive context about the environment. 
We validate \PlannerName~on real-world floorplan data, evaluated against baseline methods and human users.
An overview of this work is shown in Fig.~\ref{fig:splash}.

Our contributions are as follows:
\begin{itemize}
    \item We present insights on human performance for indoor exploration through a user study that illustrates the gap in performance to existing methods and extract actionable insights on how humans explore.
    \item We develop a new RL-based exploration planner, \PlannerName, that leverages the insights from our user study on observation space features and action spaces.
    \item Finally, we conduct thorough comparisons of SOTA frontier-based and learning-based methods, as well as human performance, to show the benefits of \PlannerName. 
\end{itemize}

\section{Related Work}\label{sec:related-works}
\subsection{Robot Exploration}
Autonomous robot exploration has been studied from multiple perspectives. Frontier-based exploration \cite{yamauchi1997frontier, best2023multi} guides robots toward the boundary between known and unknown space but often suffers from slow progress and a lack of global planning. Information-theoretic approaches seek to maximize the information gain of consecutive actions \cite{bourgault2002information, bai2016information}, yet they are often constrained by their greedy and myopic nature. Sampling-based methods \cite{nbv} construct a randomly generated tree or graph structure, such as \cite{rrt}, to expand into a traversable space, typically selecting the most informative nodes or viewpoints for exploration.

\subsection{Map Prediction for Exploration}
A common limitation of the previously mentioned approaches is their reliance on sensor data with a limited perceptual range, often failing to reason beyond the observed areas. Recently, a growing body of work has explored leveraging deep learning models to predict maps and use these predictions to enhance exploration \cite{shrestha2019mappred, luperto2021exploration, luperto2022indoorpred, saroya2020topological, ramakrishnan2020occupancy, georgakis2022upen, zwecher2022integrating, ericson2024beyondfrontier, ho_kim2024mapex}. These approaches utilize accumulated observations in a 2D top-down view as input and infer unexplored areas using predictive models trained on offline datasets. Using these predictions, robots can prioritize areas with high uncertainty \cite{georgakis2022upen}, areas with high expected coverage gain \cite{shrestha2019mappred, luperto2021exploration, luperto2022indoorpred, ericson2024beyondfrontier}, or areas with both \cite{ho_kim2024mapex}.

These approaches allow robots to reason beyond observed areas and focus on exploring areas guided by predictions; however, they still struggle due to their action selection strategy, particularly when handling environments of varying sizes and topological complexities or operating under budget constraints. In this work, we aim to address these limitations.

\subsection{Learning-based Exploration}
Studies have addressed exploration problems using deep reinforcement learning \cite{niroui2019deep, lee2021extendable, cao2023ariadne, zhu2018deep, li2019deep, chen2019self, chen2020autonomous}.  Some have focused on training policies that generate simple, primitive actions, such as moving to adjacent cells or taking a few steps in a diagonally connected grid \cite{li2019deep, chen2019self}. However, this approach often results in myopic decision-making, prioritizing immediate actions over long-term strategy.

The work in \cite{niroui2019deep} used an RL method that outputs a single weight to balance frontier scoring based on distance and estimated information gain, selecting the highest-scoring frontier. 
While this approach enables exploration of large, complex topologies, it does not directly score frontiers or exploit the predictability of indoor structures.
Some studies have employed graph structures to encode the states, defining actions as movements toward nearby nodes within the graphs \cite{lee2021extendable, cao2023ariadne, zhu2018deep}. In our work, we use frontier points as the action space and score them directly, allowing us to consider the full frontier context and avoid myopic decision-making.

Lastly, some studies have integrated the map prediction paradigm into RL-based exploration to expand the observation space \cite{zwecher2022integrating, tao2024learnexplore}. However, these studies commonly have limitations in that they set fixed primitive motions as the action space or validate their methods only on small, simplified maps. 
In our work, we expand the observation space using map prediction while simultaneously training a policy that outputs longer-horizon decision-making.

Our method, \PlannerName, differs from previous approaches in three key ways: (1) it expands the observation space beyond direct observations using map predictions, (2) it leverages RL-based methods to enhance adaptability to various human-inspired temporal and spatial contexts, 
and (3) it sets frontiers as the action space, enabling longer-horizon action selection compared to primitive motions.

\section{Problem Statement}\label{sec:prob}

We address the problem of efficient indoor mapping and exploration.
Our simulation environment is a 2D space $\mathcal{E}\subset\mathcal{R}^2$ where the robot's state is represented as a discrete 2D position $\mathbf{x}=[x, y]$. The robot is equipped with a 2D LiDAR that provides $360^\circ$ coverage to construct an occupancy grid map $O_t$ (observed map), categorizing the space as free, unknown, or occupied. At each timestep $t$, the robot moves one step $\Delta x$. We assume that the range measurements are noise-free and the robot pose is known. We aim to plan paths that maximize environment understanding as quickly as possible within a distance budget of $B$.

We define the predicted map's Intersection over Union (IoU) as a measure of exploration quality.
The exploration continues until the distance budget $B$ is reached or a sufficient understanding of the environment is achieved, which we define as 95\% IoU for the predicted map.

\section{User Study on Indoor Exploration}\label{sec:user}

We conducted a user study to uncover key human exploration strategies and compare the performance of a state-of-the-art (SOTA) algorithm MapEx \cite{ho_kim2024mapex} with human decision-making in exploring large indoor buildings.
This section covers the design and results of the user study and how we leveraged its insights to inform the design of our approach. 
The study aims to answer the following questions:\\
\textbf{Q1:} What is the performance gap between human decision-making and the SOTA exploration method, MapEx \cite{ho_kim2024mapex}? \\
\textbf{Q2:} How much does prior exposure to the map improve human exploration efficiency?\\
\textbf{Q3:} What key context do humans use during exploration?\\
\textbf{Q4:} What long-term strategies do humans use to explore?

\begin{figure}[t]
    \vspace{0.08cm}
    \centering
    \includegraphics[width=0.99\linewidth]{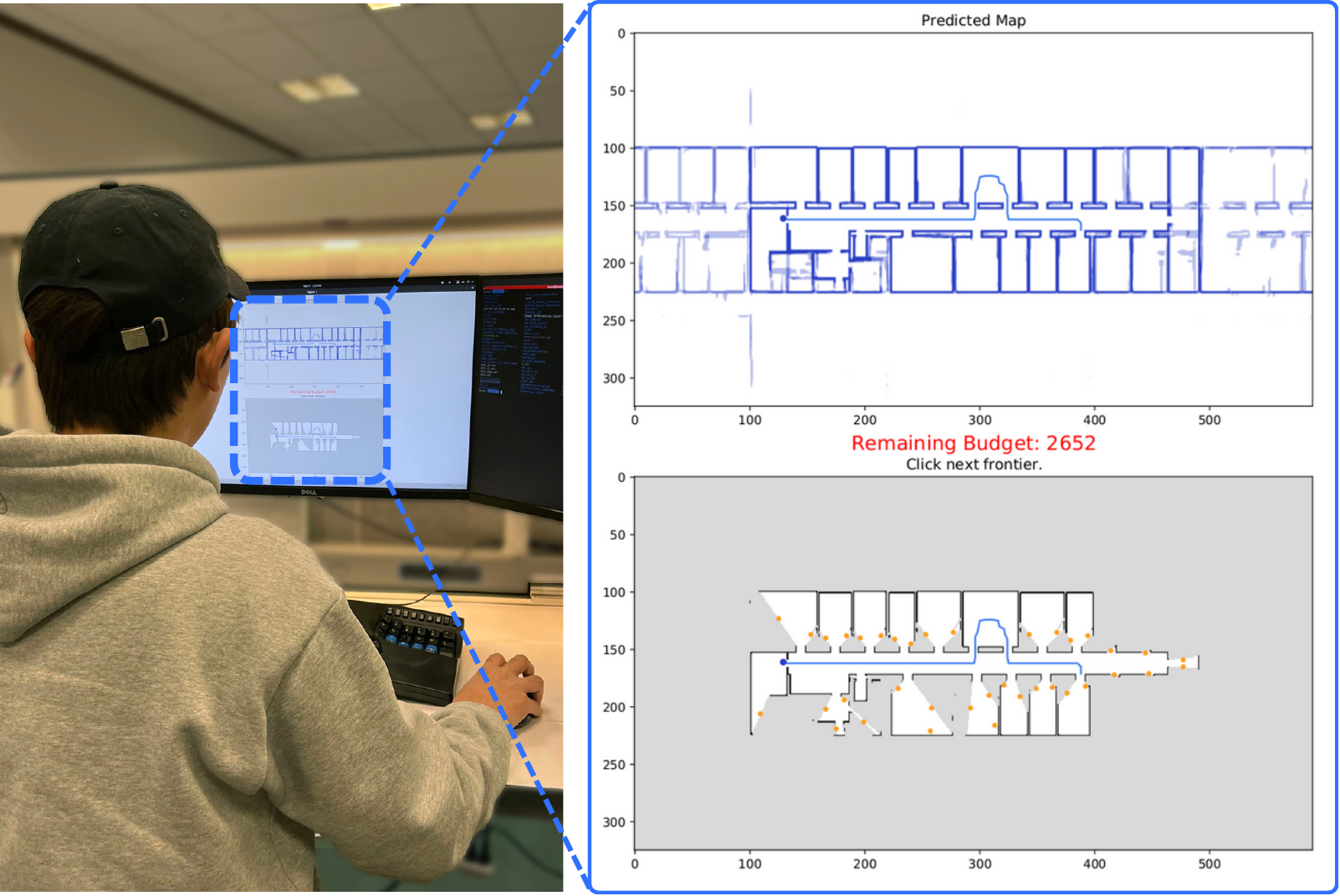}
    \caption{User interface for the study: Participants are shown the observed occupancy grid (bottom) and predicted maps (top), and must select the next frontier to explore among the frontiers marked as orange dots. The goal is to reach 95\% IoU as efficiently as possible within the distance budget $B$.}
    \label{fig:user-interface}
    \vspace{-0.25cm}
\end{figure}

\subsection{User Study Design}
To answer the above questions, we conducted a user study with 13 participants of varying experience levels in robotic exploration. To ensure a fair evaluation, participants were not allowed to see the maps beforehand, preventing any advantage from prior knowledge or mental priors. This ensured that their performance was based solely on their ability to explore the environment without prior familiarity. Users are given the same problem formulation as defined in Section~\ref{sec:prob}---either achieve 95\% IoU on the predicted map as quickly as possible or maximize IoU within the distance budget. The interface (Fig.~\ref{fig:user-interface}) displays the average output of the ensemble global map prediction network (same as \cite{ho_kim2024mapex}) and the accumulated observed occupancy grid map $O_t$. Frontiers are displayed on the observed map, and the user chooses which frontier to explore next. The robot then moves to the selected frontier, updating the maps after each robot step. This process repeats until the episode ends.

\begin{figure}[t]
    \centering
    \includegraphics[width=\columnwidth]{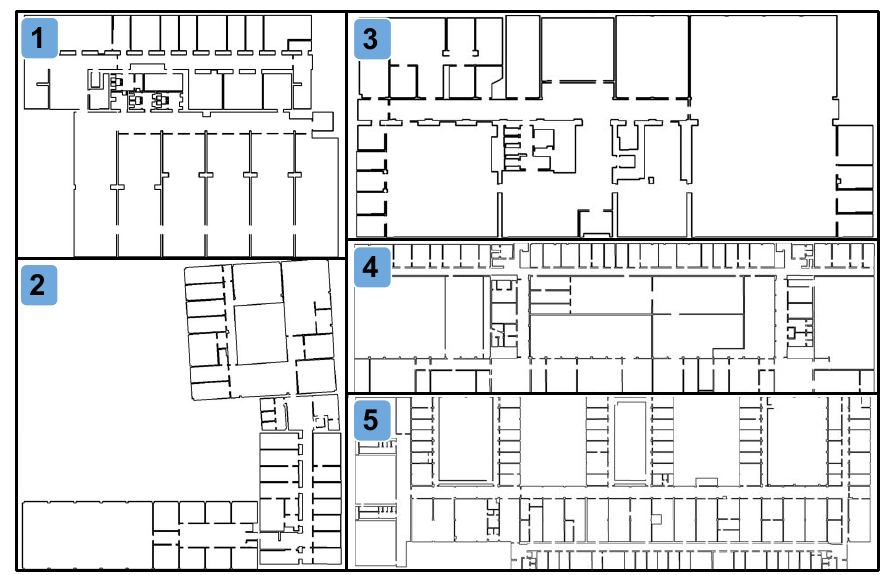}
    \caption{The set of test maps used in this work. These maps were selected to ensure a diverse range of common floorplan layouts and topologies. Note that the maps are not shown to scale relative to one another, and Map 5 is the largest in actual size, followed by Map 4.}
    \label{fig:maps-figure}
    \vspace{-0.25cm}
\end{figure}

Each participant first completes a training round to familiarize themselves with the interface and task. They then complete nine test episodes across three different maps, completing three episodes per map before moving to the next. We selected maps 1, 4, and 5 (Fig.~\ref{fig:maps-figure}) to ensure a range of map sizes and shapes. Successive episodes on the same map are referred to as rounds 1, 2, and 3. In rounds 1 and 2, participants start from the same position, while round 3 uses a different starting location. 
Round 1 assesses performance without prior knowledge of the map scale or features. Round 2 evaluates improvements based on prior knowledge and the opportunity to adapt their previous strategy. Round 3 examines how strategies adapt to new starting conditions while retaining prior map knowledge. Multiple runs per map also help capture participants' best performances. To introduce variety while maintaining comparability, we changed the set of starting positions for every three participants. Testing took approximately 30 minutes, and participants could withdraw at any point to prevent fatigue-related bias.

\subsection{Quantitative Results}
The user study results are summarized in Table~\ref{tab:user-study}, showing the average reward and 95\% confidence interval. The reward was calculated as  $R=\text{IoU}\times 1000 + B_r$, where $R$ is the reward and $B_r$ is the remaining budget at the end of the episode. MapEx results are divided based on the changes in starting positions between rounds 2 and 3. Notably, MapEx has large confidence intervals due to its deterministic nature; unlike human participants, it could not repeat runs from the same starting location.

The combined scores in the rightmost column of Table~\ref{tab:user-study} show that in round 1, even without prior knowledge of the map, participants already outperform the SOTA baseline. Examining the scores more closely, participants surpassed the baseline across all maps and rounds, except for Map 2 in round 2 and Map 4 in round 3. 
Notably, user performance on Map 2 did not improve between rounds 1 and 2. Additionally, the confidence interval for this map was significantly larger than for the others. This discrepancy may stem from the difficulty of discovering the bottom left region, which was easily overlooked. Many participants missed these rooms and continued searching elsewhere, while those who found the region achieved much higher rewards. This led to greater variance in the results, highlighting the need for reasoning beyond directly observed areas.

Generally, we observe a performance gap between humans and the SOTA exploration method. While this gap is small in round 1, it generally increases across rounds, indicating that exposure to the maps aids in exploration. Notably, performance varied among participants. Most were graduate students in robotics or related fields, with some conducting planning research and field tests using frontier-based planning methods.
These participants consistently outperformed others across all maps and rounds. One participant with extensive hands-on systems experience with robot exploration scored 907, 1220, and 1089 in round 1, and 873, 1260, and 1056 in round 3---significantly higher than peers or the baseline. This suggests that certain strategies or heuristics are more effective and that general planning experience may be beneficial. These higher scores further highlight the disparity between SOTA performance and expert human behavior.

\definecolor{tabfirst}{rgb}{1, 1, 1} %
\definecolor{tabsecond}{rgb}{1, 1, 1} %
\definecolor{tabthird}{rgb}{1, 1, 1} %

\begin{table}[t]
    \centering
    \resizebox{\linewidth}{!}{
    \begin{tabular}{lllll}\toprule
    & \textbf{Map 1} & \textbf{Map 2} & \textbf{Map 4} & \textbf{Combined} \\ \midrule
    Human Round 1  &  \cellcolor{tabthird}$1156\pm32$ & \cellcolor{tabsecond}$\mathbf{902\pm78}$ & \cellcolor{tabthird}$768\pm37$ & \cellcolor{tabthird}$911\pm60$  \\
    Human Round 2  &  \cellcolor{tabsecond}$\mathbf{1160\pm39}$ & \cellcolor{tabthird}$885\pm67$ & \cellcolor{tabsecond}$\mathbf{795\pm32}$ & \cellcolor{tabsecond}$\mathbf{915\pm56}$  \\
    MapEx Round 1\&2 &  $1071\pm770$ & $888\pm446$ & $731\pm103$ & $875\pm146$ \\
    \midrule
    Human Round 3  &  \cellcolor{tabfirst}$\mathbf{1189\pm35}$ & \cellcolor{tabfirst}$\mathbf{941\pm61}$ & \cellcolor{tabfirst}${814\pm50}$ & \cellcolor{tabfirst}$\mathbf{951\pm57 }$ \\
    MapEx Round 3 &  $1037\pm1036$ & $899\pm467$ & $\mathbf{875\pm215}$ & $924\pm116$ \\
    \bottomrule
    \end{tabular}
    }
    \caption{Results of the user study compared to MapEx \cite{ho_kim2024mapex}. Values reflect the average final reward (higher is better) and the 95\% confidence interval.}
    \label{tab:user-study}
    \vspace{-0.15cm}
\end{table}

\subsection{Qualitative Results and Insights} 

Observing the participants' strategies, those who performed best focused on maximizing prediction IoU rather than defaulting to maximizing map coverage. Some took unnecessary actions, such as searching rooms that already had clear predictions. Users often adjusted their strategies based on the map's scale---exploring rooms more in smaller maps and focusing on main hallways in larger ones. Participants also aimed to minimize backtracking, describing different approaches of thinking a few steps ahead or considering map topology for efficient traversal. The benefit of an action was often estimated by imagining the observations at the frontier. Most ignored the remaining budget until later in the episodes, when they began estimating the cost of reaching frontiers with limited steps left. Many were surprised at the task's difficulty and reported feeling fatigued by the end.

\begin{figure}[t]
    \centering
    \includegraphics[trim={0.5cm 0cm 0cm 0cm},clip,width=0.99\linewidth]{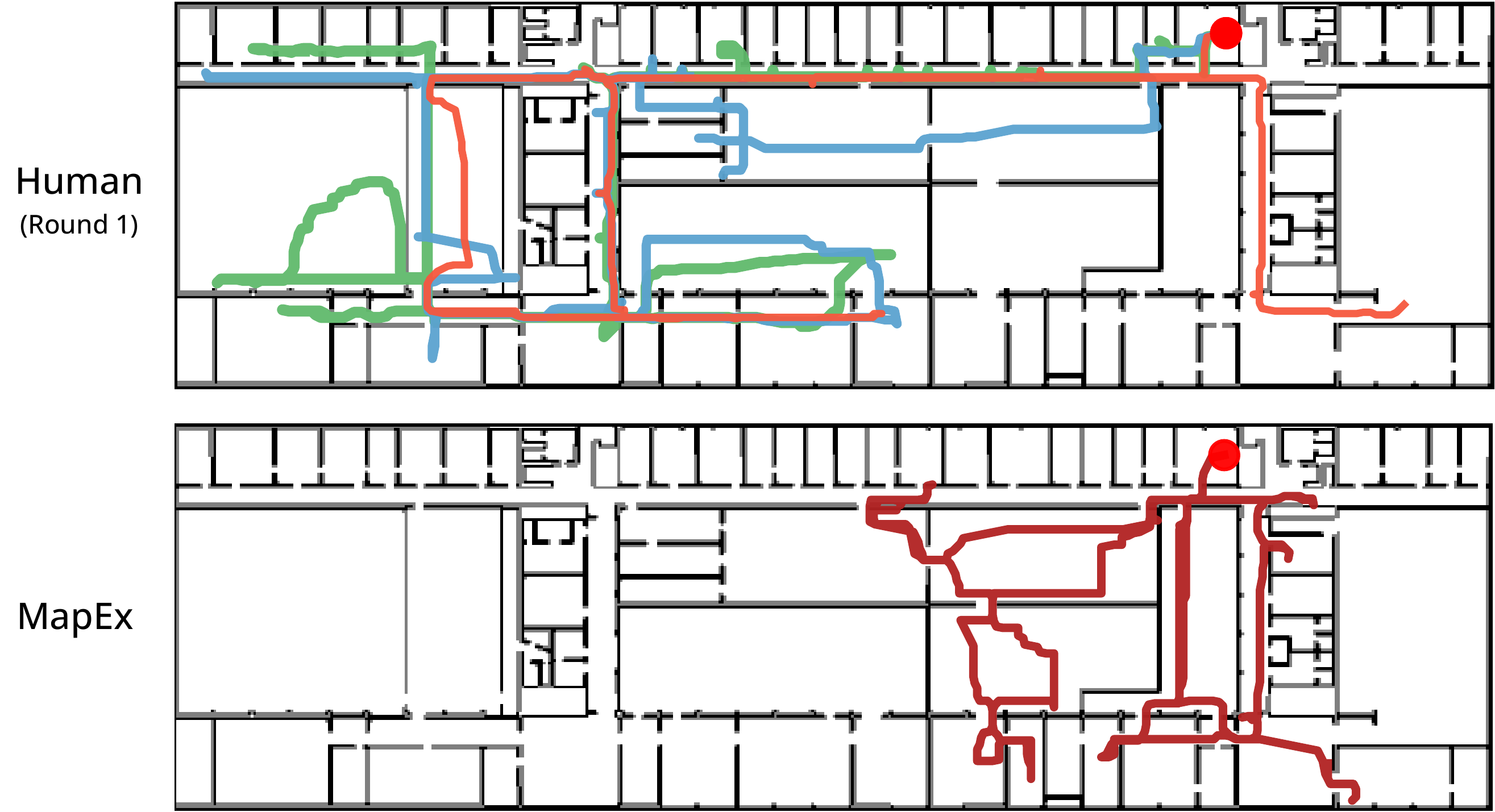}
    \caption{Three user runs and MapEx on Map 4, with colors differentiating the separate users and the red dot marking the starting position. Users explored more evenly across the space, leading to higher final reward. }
    \label{fig:user-visuals}
    \vspace{-0.25cm}
\end{figure}

Fig.~\ref{fig:user-visuals} shows the paths chosen by three users in round 1 compared to MapEx. MapEx concentrated its exploration budget in a limited region, resulting in insufficient exploration of the entire space. In contrast, the human users distributed their exploration more evenly, despite having no prior knowledge of the environment. We observed that between rounds 1 and 2, where the start position remained the same, users realized the overall size and shape of the space and adjusted their strategies to explore more evenly.

From the quantitative and qualitative results from the user experiments, we found the following actionable insights:
\begin{enumerate}
    \item Leveraging the global map predictions can likely benefit exploration as it allows for maximization of prediction accuracy through prioritizing observing uncertain areas in the predictions instead of exploring exhaustively. \label{finding1}
    \item The value of an action relates to the balance of exploitation and exploration while also considering both area coverage and prediction uncertainty. \label{finding2}
    \item Strategies must adapt to the map scale relative to the budget, as a single heuristic cannot perform effectively across all map sizes. \label{finding3}
    \item An ideal policy has budget awareness throughout the entire planning process to minimize backtracking. Thinking more than one step ahead helps avoid backtracking and creates more efficient paths. \label{finding4}
\end{enumerate}

\section{Approach}\label{sec:approach}
\begin{figure*}[ht]
    \centering
\includegraphics[trim={0.0cm 0.0cm 0.0cm 0.0cm},clip,width=0.98\linewidth]{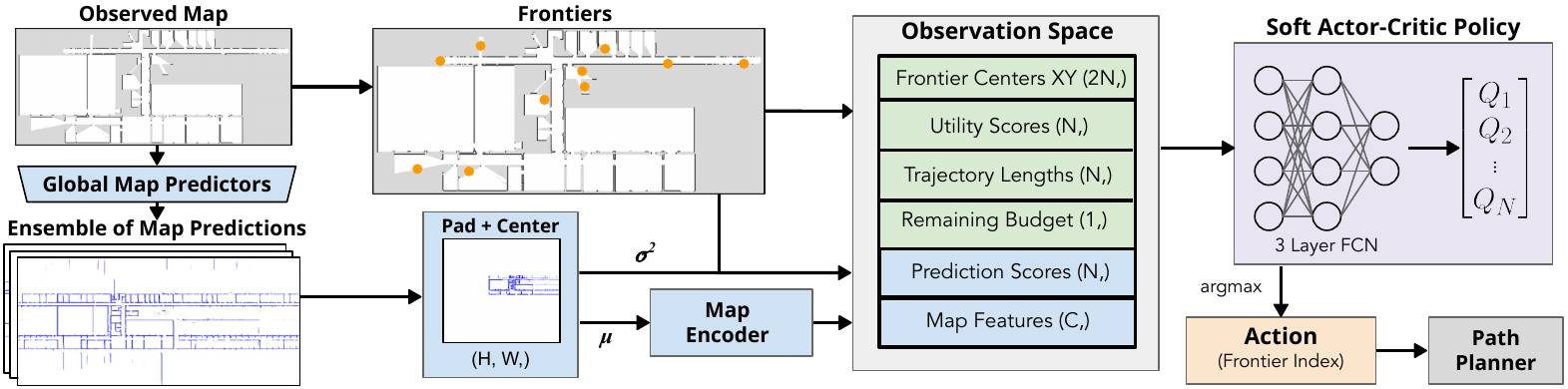}    
\caption{The \PlannerName~pipeline, where observed maps are processed through three independent global map prediction models, generating prediction maps. These maps are averaged and passed through a convolutional encoder to extract a 256-dimensional feature vector. This vector is concatenated with frontier centers, prediction and utility scores, distances from the agent, and remaining budget. The resulting vector is fed into a fully connected network that outputs $N$ values, and the argmax is selected as the index of the frontier action.}
    \label{fig:approach-pipeline}
\end{figure*}

Based on our user study, we propose a Reinforcement Learning (RL) pipeline using the Soft Actor-Critic (SAC) algorithm \cite{haarnoja2018soft}, leveraging frontiers as an efficient action space. Our policy input incorporates global map predictions for global context and calculates predicted frontier information gain. This approach enables the model to integrate multiple scoring methods (user insights \ref{finding1} and \ref{finding2}), global context (user insight \ref{finding3}), and remaining budget (user insight \ref{finding4}) to select actions more effectively than heuristic-based methods. Fig.~\ref{fig:approach-pipeline} provides an overview. For implementation, we used Stable Baselines 3 \cite{stable-baselines3}, with SAC consisting of two fully connected layers for both actor and critic.

\subsection{Map Encoder}
Building on user insight \ref{finding1}, we use the map prediction model \cite{LAMA}, which predicts the global map based on the accumulated occupancy map from LiDAR observations. Prior work \cite{ho_kim2024mapex} has shown its effectiveness in estimating global topology beyond the robot's current observations. Following \cite{georgakis2022upen, ho_kim2024mapex}, we model the prediction uncertainty by generating multiple independent predictions from an ensemble of networks and calculating the variance of their outputs. This encourages a robot to prioritize uncertain areas.

Using the average of the ensemble of predicted maps, we generate a top-down map centered on the robot's position, applying padding and cropping to produce a 1600x1600 pixel map. This map is then resized to 256x256 pixels using interpolation, followed by max pooling to obtain a 128x128 pixel input for the encoder. These operations ensure consistency across maps of varying shapes and sizes, following user insight \ref{finding3}---no single heuristic works optimally in all scenarios, and strategies must adjust to different map scales.

The 128×128 pixel map is processed by an encoder with three convolutional layers and a fully connected layer, producing a $C$=256 dimensional latent vector. Separate map encoders are used for the actor and critic.

\subsection{Observation and Action Spaces}
When a new action needs to be chosen, frontiers are calculated from the current occupancy map. In order to have a fixed action space size, this set of frontiers must be reduced to a fixed number $N$.
We choose the top $N$ frontiers based on their prediction scores, excluding those outside of the robot-centered map. In cases when there are fewer than $N$ frontier centers, we add zero padding for the remaining. In our implementation, we select $N=10$ frontiers based on experimental results, as they provide sufficient exploration options in all directions across training and testing maps.

To ensure the selected frontiers are unique and well-distributed across the map, we compare the prediction and utility scores of frontier points within a 5-meter distance threshold. If the score differences are below an experimentally determined threshold (0.01) and the points are spatially close, we remove duplicates to maintain diversity in the observation space.

The observation space includes the 256 encoded map features, $[x, y]$ coordinates of the $N$ frontiers, utility and prediction scores for these frontiers, trajectory length from the robot's current position to each frontier, and the remaining distance budget.
Each frontier center is defined relative to the robot's position to align with the map transform. 

The frontier features are calculated as follows:
\begin{itemize} 
    \item Frontier Centers: convolve an edge detection kernel across the observed map, group and filter to be above a minimum frontier edge size (following \cite{yamauchi1997frontier}), find the center, translate into a robot-centered convention, and normalize by half the robot-centered map width.
    \item Utility Scores: raycast and create visibility mask in the occupancy map, sum the number of unknown cells in the visibility mask, normalize by the A* distance to the frontier, and apply min-max normalization. 
    \item Prediction Scores: probabilistic raycast in the mean predicted map, sum the variance of the cells in the visibility mask, normalize by the A* distance (rather than Euclidean as in \cite{ho_kim2024mapex}), and apply min-max normalization.
    \item Trajectory Lengths: A* distances to the frontiers from the robot's current position, normalized by $B$
    \item Remaining Budget $B_r$: the remaining mission budget, normalized by the total budget $B$.
\end{itemize}

The policy network outputs a $N$-dimensional vector (action space), where each value represents one of the top $N$ frontier options. 
The frontier corresponding to the highest value in this vector is selected as the next action.

\begin{table*}[th]
\centering
\resizebox{\textwidth}{!}{
\begin{tabular}{lm{0.9cm}m{1.3cm}m{0.9cm}m{1.3cm}m{0.9cm}m{1.3cm}m{0.9cm}m{1.3cm}m{0.9cm}m{1.3cm}m{0.9cm}m{1.3cm}}
\toprule
 & \multicolumn{2}{c}{\textbf{Map 1}} & \multicolumn{2}{c}{\textbf{Map 2}} & \multicolumn{2}{c}{\textbf{Map 3}} & \multicolumn{2}{c}{\textbf{Map 4}} & \multicolumn{2}{c}{\textbf{Map 5}} & \multicolumn{2}{c}{\textbf{Combined}} \\
\cline{2-13}
 \textbf{Method} & Reward & IoU & Reward & IoU & Reward & IoU & Reward & IoU & Reward & IoU & Reward & IoU \\
\cmidrule(lr){1-1}\cmidrule(lr){2-3}\cmidrule(lr){4-5}\cmidrule(lr){6-7}\cmidrule(lr){8-9}\cmidrule(lr){10-11}\cmidrule(lr){12-13}
\textbf{\PlannerName~(ours)} & \bf 726\,$\pm$\,50 & \bf 0.96\,$\pm$\,0.01 & \bf 465\,$\pm$\,91 & \bf 0.84\,$\pm$\,0.07 & \bf 722\,$\pm$\,41 & \bf 0.96\,$\pm$\,0.00 & 370\,$\pm$\,40 & 0.77\,$\pm$\,0.04 & \bf 329\,$\pm$\,27 & \bf 0.73\,$\pm$\,0.03 & \bf 522\,$\pm$\,46 & \bf 0.85\,$\pm$\,0.03 \\
MapEx \cite{ho_kim2024mapex} & 715\,$\pm$\,40 & 0.96\,$\pm$\,0.00 & 404\,$\pm$\,90 & 0.79\,$\pm$\,0.07 & 695\,$\pm$\,52 & 0.95\,$\pm$\,0.00 & \bf 397\,$\pm$\,\bf36 & \bf0.80\,$\pm$\,0.04 & 277\,$\pm$\,40 & 0.68\,$\pm$\,0.04 & 498\,$\pm$\,48 & 0.83\,$\pm$\,0.03 \\
\PlannerName~No Pred & 723\,$\pm$\,69 & 0.95\,$\pm$\,0.01 & 386\,$\pm$\,94 & 0.77\,$\pm$\,0.07 & 677\,$\pm$\,49 & 0.95\,$\pm$\,0.00 & 368\,$\pm$\,31 & 0.76\,$\pm$\,0.03 & 314\,$\pm$\,38 & 0.71\,$\pm$\,0.03 & 494\,$\pm$\,47 & 0.83\,$\pm$\,0.02 \\
Random Policy & 562\,$\pm$\,88 & 0.90\,$\pm$\,0.04 & 384\,$\pm$\,57 & 0.78\,$\pm$\,0.06 & 577\,$\pm$\,52 & 0.94\,$\pm$\,0.01 & 266\,$\pm$\,23 & 0.67\,$\pm$\,0.02 & 232$\pm$32 & 0.63\,$\pm$\,0.03 & 401\,$\pm$\,41 & 0.78\,$\pm$\,0.03 \\
Nearest Frontier & 482\,$\pm$\,15 & 0.88\,$\pm$\,0.02 & 310\,$\pm$\,33 & 0.65\,$\pm$\,0.08 & 561\,$\pm$\,67 & 0.90\,$\pm$\,0.04 & 232\,$\pm$\,27 & 0.63\,$\pm$\,0.03 & 94\,$\pm$\,47 & 0.49\,$\pm$\,0.05 & 336\,$\pm$\,44 & 0.71\,$\pm$\,0.04 \\
RL Motion Primitive & 57\,$\pm$\,36 & 0.45\,$\pm$\,0.04 & 16\,$\pm$\,23 & 0.33\,$\pm$\,0.06 & 77\,$\pm$\,40 & 0.45\,$\pm$\,0.06 & 0\,$\pm$\,0 & 0.29\,$\pm$\,0.05 & 0\,$\pm$\,0 & 0.20$\pm$0.07 & 30\,$\pm$\,13 & 0.34\,$\pm$\,0.03 \\
\bottomrule
\end{tabular}
}
\caption{Average and 95\% confidence interval for reward and IoU on the test maps. Each map includes 15 experiments from different starting positions, totaling 75 experiments across 5 maps.}
\label{table:policy_evaluations}
\vspace{-0.25cm}
\end{table*}

\subsection{Reward Function} 
To achieve 95\% IoU as efficiently as possible, our reward function considers both the final episode IoU and the remaining budget. 
We designed the reward function to provide feedback only at the end of the episode, based on the accuracy of the predicted map compared to the ground truth. This fully sparse reward is feasible given our efficient action space, which inherently limits the number of steps per episode. The reward $R$ is defined as
\begin{equation*}\label{equ:reward}
\scalebox{0.9}{$ %
R = 
\begin{cases} 
\max(0, \text{IoU}-\delta_{\text{IoU}}) \times 1000 + B_r, & B_r = 0 \text{ or } \text{IoU} \geq 0.95 \\ 
0, & \text{otherwise}
\end{cases}
$}
\end{equation*}

where $B_r$ represents the remaining mission time or budget, while $\delta_{\text{IoU}}=0.4$ clips the IoU to create a more evenly distributed reward signal and strengthen the feedback to the network. With nearly all training episodes achieving an IoU above $0.4$, this clipping significantly improved training performance. This reward function encourages efficient exploration by rewarding the model for reaching 95\% IoU while minimizing total distance usage.

We use the Intersection over Union (IoU) of the occupied cells for the global map prediction as our metric of performance. IoU is computed as $\text{IoU} = \frac{\text{TP}}{\text{FP} + \text{TN} + \text{TP}}$, 
where $\text{TP}, \text{FP},\text{TN}$ denote true positives, false positives, and true negatives, respectively. To avoid penalizing minor map discrepancies---like wall thickness variations, gaps, dataset artifacts, or small unobservable spaces---we use binary dilations as a buffer in IoU calculation.

Let $P$ represent the predicted binary map of occupied cells, $G$ represent the ground truth occupied binary map, and $O$ represent the observed map. 
We define the masked prediction $P_m$ such that $P_m = P$ where $O = 0.5$ (unknown space) and $P_m = O$ otherwise, where only unknown space in the observed map is replaced by the global map prediction. We then compute a corrected prediction $P_c$ from $P_m$ by excluding all predictions that fall outside the building. We dilate $P_m$ and $G$ by a fixed kernel size of 5 pixels (50 cm) to create $P_d$ and $G_d$, respectively. By evaluating all pixels $i$, we can then calculate $\text{TP} = \sum_{i} (P_d(i) \land G(i))$, $\text{FP} = \sum_{i} (P_m(i) \land P_c(i) \land \neg G_d(i))$, and $\text{FN} = \sum_{i} (\neg P_d(i) \land G(i))$,
where $\land$ and $\neg$ denote the logical AND and NOT operators, respectively. This process allows the IoU to focus on overall map topology and wall placement. Without this correction, IoU plateaus at 75-85\%, even with full area coverage.

\subsection{Overall Exploration Cycle} 
At the start of each episode, the robot continuously builds a 2D occupancy map with the incoming LiDAR observations. The observed map is then fed into 3 independent global map prediction models, each trained on different subsets of the training maps, to produce an average map and variance map for frontier's prediction score calculations.
After selecting the top $N$ frontiers, the frontier features and the mean map are computed and subsequently provided to the policy model, which determines the frontier for the robot to explore.
An A* path from the current location to the frontier center is found, and the robot advances until the goal is reached or is deemed unreachable.
This iterative process persists until the mission budget $B$ or the robot successfully predicts the global occupancy map with 95\% IoU.

\subsection{Simulation Details}
We evaluate \PlannerName~on the real-world KTH floorplan dataset \cite{aydemir2012kthdataset}.
The dataset consists of 149 campus floorplan blueprints from the KTH Royal Institute of Technology, which detail wall and door positions. We use 7 maps for training and 5 maps (shown in Fig.~\ref{fig:maps-figure}) for testing \PlannerName, selecting maps that are medium to large in size, have complex topologies, and were not in the training set for the global map predictor model. 
We simulate a robot operating in the building with a LiDAR sensor that has 2500 scanning beams at a 20-meter range. Our maps are scaled such that one pixel is equal to 10 cm.

\begin{figure*}[ht]
    \centering
    \includegraphics[width=\textwidth]{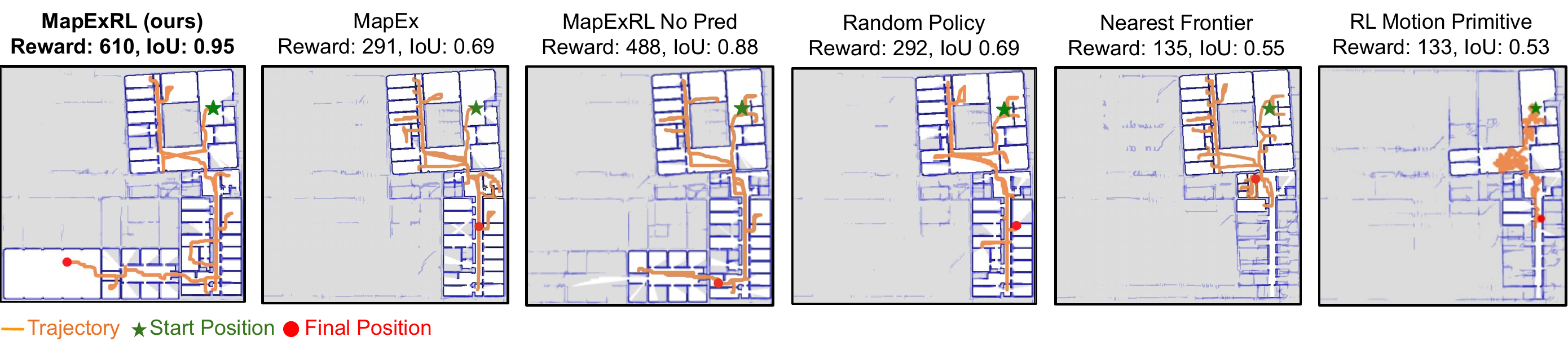}    
    \caption{Comparison of the six evaluated methods on Map 2 from the same start position. Notably, \PlannerName~is the only one that successfully explored the entire building and navigated the narrow passage leading to the bottom-left section that was not reached by the baseline methods.}
    \label{fig:example-outputs}
    \vspace{-0.25cm}
\end{figure*}

\subsection{Baselines} 
Our first baseline is the conventional frontier-based exploration method \cite{yamauchi1997frontier}, where the agent chooses the nearest frontier based on Euclidean distance, without utilizing any global map predictions. Our second baseline, MapEx \cite{ho_kim2024mapex}, selects the frontier with the highest information gain, calculated by jointly considering sensor coverage and variance from predicted maps, normalized by the Euclidean distance. We also include a random exploration baseline, where the agent randomly selects a frontier from the top $N$ frontiers in the observation space. This baseline validates that our model learns to navigate efficiently using the top $N$ frontiers rather than randomly picking from the list. 

Additionally, we compare against a baseline inspired by \cite{zwecher2022integrating}, which uses RL with motion primitives as actions. We adopt their action spaces and reward functions to train the policy while employing our global map prediction module. The action space consists of 8 primitive motions ($\mathcal{A} = { \nwarrow, \uparrow, \nearrow, \leftarrow, \rightarrow, \swarrow, \downarrow, \searrow }$), each with a constant step size $\Delta x$, uniform across all methods to ensure consistency with our approach. The reward function heavily penalizes actions that result in collisions with occupied cells and rewards actions that increase the number of newly observed cells. For the details of the reward function, refer to Equation 4 of \cite{zwecher2022integrating}. This baseline helps validate the impact of using frontier points as action spaces instead of primitive motions. We used the PPO algorithm to train the policy, following the approach in \cite{zwecher2022integrating}, and trained the model with the same number of steps to ensure a fair comparison with our method.

Lastly, we compared our model with an ablation model, referred to as \PlannerName~No Pred, which excludes global map predictions from its observation space. If our model outperforms this ablation, it validates user insight \ref{finding1}, confirming the importance of incorporating global map predictions.

\subsection{Training Details} 
During training, we set a maximum mission budget of $B=500$, which is equivalent to 300 meters. The episode ends when 
the predicted map reaches 95\% IoU or the robot exhausts the mission time.
For comparison across all methods, we calculate IoU for each step using the predicted map, including for baselines that do not incorporate global predictions for frontier scoring. 

The discount factor for SAC is $\gamma = 0.99$. The batch size is set to 256, the replay buffer capacity is 10,000, and training begins once the replay buffer exceeds 1,000 data points. Training steps is set to 4, $\tau$ is set to 0.02, and training frequency is set to 1. The Adam optimizer is used with a learning rate of 0.00073 for both policy and critic networks. 
Our model is trained on an NVIDIA A100 GPU, which typically takes less than a day without parallelization to achieve 10,000 steps and train the optimal model.

\section{Results}\label{sec:results}

The performance of our model was evaluated on the 5 test maps with the same 15 randomly assigned start positions per map, resulting in a total of 75 experiments per method. The detailed results showing the average and 95\% confidence interval for reward and IoU are presented in Table~\ref{table:policy_evaluations}.

Our policy outperforms all baseline methods in four of five test maps, ranking second only on Map 4. It also surpasses the ablation model across all maps, validating user insight \ref{finding1}, which motivated this ablation. These results demonstrate our model’s ability to reason about and effectively utilize predicted maps, exploration budget, and robot observations, reinforcing the correctness of the users' insights.
Notably, our policy performed exceptionally on Map 2 and Map 5, with Map 5 being the largest and most complex in our dataset.
For comparison, the nearest frontier-based exploration method achieves an average IoU of only 49\% on Map 5, whereas our method reaches 73\%.
This map also underscores the inherent challenges associated with large maps and emphasizes the advantages of predictive global maps. 
The nearest-frontier method exhausts its budget, exploring only a small portion of the map, due to many close-proximity frontiers.

Map 2 is an irregular L-shaped building that is challenging to explore, creating numerous opportunities for backtracking, which is a common issue with the baseline methods. In contrast, our policy often effectively selected paths that minimized backtracking and efficiently explored hard-to-predict areas using the global map prediction models.
A sample trajectory across all methods on Map 2 is illustrated in Fig.~\ref{fig:example-outputs}, demonstrating the efficiency of our policy in topologically challenging environments. 
As observed, the global map prediction models struggled to predict the bottom-left part of the building accurately, and our approach was the only method to successfully explore that area. 

Overall, we observe a 4.8\% improvement in exploration efficiency in all 75 combined evaluations compared to MapEx, the strongest baseline, 
with improvements reaching up to 18.8\% on the largest and most topologically challenging Map 5.
Additionally, we achieved a 30\% improvement in performance compared to the random policy and a 55\% improvement over the traditional nearest-frontier exploration method. 
Notably, our model outperformed all baselines on the largest and most topologically complex map (Map 5) and demonstrated strong performance on the irregular L-shaped map (Map 2), highlighting its advantages in challenging exploration tasks and its efficient use of Global Map Predictions and budget awareness. While performance gains are less pronounced on smaller maps due to their simplicity and limited time available to differentiate in performance, our model still maintains a consistent advantage.

Despite our careful implementation of the RL motion primitive method and earnest efforts to optimize the various parameters, the planner showed limited effectiveness. As shown in Fig.~\ref{fig:example-outputs}, this approach tended to resemble a random walk, with prolonged training primarily leading to a reduction in wall collisions. We verified our motion-primitive RL baseline on a simpler map, where it converged and explored well. However, when trained and tested on larger, complex maps, performance degraded significantly, likely due to many factors, including narrow doorways that are difficult to navigate and a weak training signal, as most actions yield no reward given the maps’ scale. While the model learned to avoid collisions, efficient exploration remained a challenge. This outcome highlights one of the advantages of our action space, which employs classical methods to effectively extract a set of actions from the map that will facilitate new observations.

To evaluate our method against the baseline established by our user study, we tested the \PlannerName~planner using the two sets of starting positions utilized in the study. Using the reward function from the user study, the combined reward for the start positions for rounds 1 and 2 was $860 \pm 176$, while for round 3 it was $956 \pm 164$. Notably, our model outperformed the human baseline in round 3. However, it did not surpass the users' performance in rounds 1 and 2, primarily due to a few low-reward runs that significantly impacted the average score. While a larger sample size would allow a more definitive comparison with human operators, these results underscore the continued high performance of skilled human operators and highlight the necessity for robustness and consistency in an exploration planner. Furthermore, the findings indicate that there remains room for improvement in our approach to further bridge the performance gap between robot and human exploration strategies.

\section{Conclusion}\label{sec:conclusion}
In this work, we present \PlannerName, a novel human-inspired method combining Reinforcement Learning with frontier-based exploration and global map predictions to improve indoor mapping efficiency. Our user study informs key design choices and provides a benchmark for evaluating exploration methods.
By optimizing frontier selection through a learned policy, \PlannerName~improves exploration strategies, achieving higher predicted map IoU with fewer resources. 
Experimental results on a real-world floorplan dataset shows its advantage over traditional SOTA methods and ablation studies, particularly in large, complex environments. 

In future work, we plan to extend our approach to cluttered environments and real-world robotic deployment.
We also aim to refine the problem formulation by initiating exploration with a map of the building's exterior walls, a practical assumption for many applications. This provides spatial scale awareness, enabling the model to optimize policy selection within budget constraints without requiring prior map scale discovery.

\section*{ACKNOWLEDGMENT}
The authors want to thank all the people behind the CMU RISS program, especially Rachel Burcin and Dr. John M. Dolan. 
Special thanks to Kasper Grøntved, Jeric Lew, and Graeme Best for helfpul discussions.

\bibliographystyle{ieeetr}
\bibliography{sample}

\begin{thebibliography}{10}

\bibitem{yamauchi1997frontier}
B.~Yamauchi, ``A frontier-based approach for autonomous exploration,'' in {\em Proc. of IEEE International Symposium on Computational Intelligence in Robotics and Automation (CIRA)}, pp.~146--151, 1997.

\bibitem{oriolo1998real}
G.~Oriolo, G.~Ulivi, and M.~Vendittelli, ``Real-time map building and navigation for autonomous robots in unknown environments,'' {\em IEEE Transactions on Systems, Man, and Cybernetics, Part B (Cybernetics)}, vol.~28, no.~3, pp.~316--333, 1998.

\bibitem{best2023multi}
G.~Best, R.~Garg, J.~Keller, G.~A. Hollinger, and S.~Scherer, ``Multi-robot, multi-sensor exploration of multifarious environments with full mission aerial autonomy,'' {\em The International Journal of Robotics Research}, vol.~43, no.~4, pp.~485--–512, 2024.

\bibitem{kim2023multi}
S.~Kim, M.~Corah, J.~Keller, G.~Best, and S.~Scherer, ``Multi-robot multi-room exploration with geometric cue extraction and circular decomposition,'' {\em IEEE Robotics and Automation Letters}, 2023.

\bibitem{georgakis2022upen}
G.~Georgakis, B.~Bucher, A.~Arapin, K.~Schmeckpeper, N.~Matni, and K.~Daniilidis, ``Uncertainty-driven planner for exploration and navigation,'' in {\em IEEE Conference on Robotics and Automation (ICRA)}, 2022.

\bibitem{shrestha2019mappred}
R.~Shrestha, F.-P. Tian, W.~Feng, P.~Tan, and R.~Vaughan, ``Learned map prediction for enhanced mobile robot exploration,'' in {\em Proc. of International Conference on Robotics and Automation (ICRA)}, pp.~1197--1204, 2019.

\bibitem{ho_kim2024mapex}
C.~Ho, S.~Kim, B.~Moon, A.~Parandekar, N.~Harutyunyan, C.~Wang, K.~Sycara, G.~Best, and S.~Scherer, ``Mapex: Indoor structure exploration with probabilistic information gain from global map predictions,'' in {\em 2025 IEEE International Conference on Robotics and Automation (ICRA)}, 2025.

\bibitem{niroui2019deep}
F.~Niroui, K.~Zhang, Z.~Kashino, and G.~Nejat, ``Deep reinforcement learning robot for search and rescue applications: Exploration in unknown cluttered environments,'' {\em IEEE Robotics and Automation Letters}, vol.~4, no.~2, pp.~610--617, 2019.

\bibitem{li2019deep}
H.~Li, Q.~Zhang, and D.~Zhao, ``Deep reinforcement learning-based automatic exploration for navigation in unknown environment,'' {\em IEEE transactions on neural networks and learning systems}, vol.~31, no.~6, pp.~2064--2076, 2019.

\bibitem{lee2021extendable}
W.-C. Lee, M.~C. Lim, and H.-L. Choi, ``Extendable navigation network based reinforcement learning for indoor robot exploration,'' in {\em 2021 IEEE International Conference on Robotics and Automation (ICRA)}, pp.~11508--11514, IEEE, 2021.

\bibitem{cao2023ariadne}
Y.~Cao, T.~Hou, Y.~Wang, X.~Yi, and G.~Sartoretti, ``Ariadne: A reinforcement learning approach using attention-based deep networks for exploration,'' in {\em 2023 IEEE International Conference on Robotics and Automation (ICRA)}, pp.~10219--10225, IEEE, 2023.

\bibitem{bourgault2002information}
F.~Bourgault, A.~A. Makarenko, S.~B. Williams, B.~Grocholsky, and H.~F. Durrant-Whyte, ``Information based adaptive robotic exploration,'' in {\em Proc. of IEEE/RSJ International Conference on Intelligent Robots and Systems (IROS)}, vol.~1, pp.~540--545, 2002.

\bibitem{bai2016information}
S.~Bai, J.~Wang, F.~Chen, and B.~Englot, ``Information-theoretic exploration with {B}ayesian optimization,'' in {\em Proc. of IEEE/RSJ International Conference on Intelligent Robots and Systems (IROS)}, pp.~1816--1822, 2016.

\bibitem{nbv}
A.~Bircher, M.~Kamel, K.~Alexis, H.~Oleynikova, and R.~Siegwart, ``Receding horizon" next-best-view" planner for 3d exploration,'' in {\em 2016 IEEE international conference on robotics and automation (ICRA)}, pp.~1462--1468, IEEE, 2016.

\bibitem{rrt}
S.~LaValle, ``Rapidly-exploring random trees: A new tool for path planning,'' {\em Research Report 9811}, 1998.

\bibitem{luperto2021exploration}
M.~Luperto, L.~Fochetta, and F.~Amigoni, ``Exploration of indoor environments through predicting the layout of partially observed rooms,'' in {\em Proceedings of the 20th International Conference on Autonomous Agents and MultiAgent Systems}, AAMAS '21, (Richland, SC), p.~836–843, International Foundation for Autonomous Agents and Multiagent Systems, 2021.

\bibitem{luperto2022indoorpred}
M.~Luperto and F.~Amigoni, ``Reconstruction and prediction of the layout of indoor environments from two-dimensional metric maps,'' {\em Engineering Applications of Artificial Intelligence}, vol.~113, p.~104910, 2022.

\bibitem{saroya2020topological}
M.~Saroya, G.~Best, and G.~A. Hollinger, ``Online exploration of tunnel networks leveraging topological {CNN}-based world predictions,'' in {\em Proc. of IEEE/RSJ International Conference on Intelligent Robots and Systems (IROS)}, pp.~6038--6045, 2020.

\bibitem{ramakrishnan2020occupancy}
S.~K. Ramakrishnan, Z.~Al-Halah, and K.~Grauman, ``Occupancy anticipation for efficient exploration and navigation,'' in {\em Proc. of European Conference on Computer Vision (ECCV)}, pp.~400--418, 2020.

\bibitem{zwecher2022integrating}
E.~Zwecher, E.~Iceland, S.~R. Levy, S.~Y. Hayoun, O.~Gal, and A.~Barel, ``Integrating deep reinforcement and supervised learning to expedite indoor mapping,'' in {\em 2022 International Conference on Robotics and Automation (ICRA)}, pp.~10542--10548, IEEE, 2022.

\bibitem{ericson2024beyondfrontier}
L.~Ericson and P.~Jensfelt, ``Beyond the frontier: Predicting unseen walls from occupancy grids by learning from floor plans,'' {\em IEEE Robotics and Automation Letters}, vol.~9, no.~8, pp.~6832--6839, 2024.

\bibitem{zhu2018deep}
D.~Zhu, T.~Li, D.~Ho, C.~Wang, and M.~Q.-H. Meng, ``Deep reinforcement learning supervised autonomous exploration in office environments,'' in {\em 2018 IEEE international conference on robotics and automation (ICRA)}, pp.~7548--7555, IEEE, 2018.

\bibitem{chen2019self}
F.~Chen, S.~Bai, T.~Shan, and B.~Englot, ``Self-learning exploration and mapping for mobile robots via deep reinforcement learning,'' in {\em Aiaa scitech 2019 forum}, p.~0396, 2019.

\bibitem{chen2020autonomous}
F.~Chen, J.~D. Martin, Y.~Huang, J.~Wang, and B.~Englot, ``Autonomous exploration under uncertainty via deep reinforcement learning on graphs,'' in {\em 2020 IEEE/RSJ International Conference on Intelligent Robots and Systems (IROS)}, pp.~6140--6147, IEEE, 2020.

\bibitem{tao2024learnexplore}
Y.~Tao, E.~Iceland, B.~Li, E.~Zwecher, U.~Heinemann, A.~Cohen, A.~Avni, O.~Gal, A.~Barel, and V.~Kumar, ``Learning to explore indoor environments using autonomous micro aerial vehicles,'' in {\em 2024 IEEE International Conference on Robotics and Automation (ICRA)}, pp.~15758--15764, 2024.

\bibitem{haarnoja2018soft}
T.~Haarnoja, A.~Zhou, K.~Hartikainen, G.~Tucker, S.~Ha, J.~Tan, V.~Kumar, H.~Zhu, A.~Gupta, P.~Abbeel, {\em et~al.}, ``Soft actor-critic algorithms and applications,'' {\em arXiv preprint arXiv:1812.05905}, 2018.

\bibitem{stable-baselines3}
A.~Raffin, A.~Hill, A.~Gleave, A.~Kanervisto, M.~Ernestus, and N.~Dormann, ``Stable-baselines3: Reliable reinforcement learning implementations,'' {\em Journal of Machine Learning Research}, vol.~22, no.~268, pp.~1--8, 2021.

\bibitem{LAMA}
R.~Suvorov, E.~Logacheva, A.~Mashikhin, A.~Remizova, A.~Ashukha, A.~Silvestrov, N.~Kong, H.~Goka, K.~Park, and V.~Lempitsky, ``Resolution-robust large mask inpainting with {F}ourier convolutions,'' in {\em Proc. of IEEE/CVF Winter Conference on Applications of Computer Vision}, pp.~2149--2159, 2022.

\bibitem{aydemir2012kthdataset}
A.~Aydemir, P.~Jensfelt, and J.~Folkesson, ``What can we learn from 38,000 rooms? {R}easoning about unexplored space in indoor environments,'' in {\em Proc. of IEEE/RSJ International Conference on Intelligent Robots and Systems (IROS)}, pp.~4675--4682, 2012.

\end{thebibliography}

\end{document}